\documentclass{bmvc2k}

\title{SymmNet: A Symmetric Convolutional Neural Network for Occlusion Detection}

\addauthor{Ang Li}{bennie.522@stu.xjtu.edu.cn}{1}
\addauthor{Zejian Yuan}{yuan.ze.jian@xjtu.edu.cn}{1}

\addinstitution{
 Institute of Artificial Intelligence and Robotics\\
 Xi'an Jiaotong University\\
 Xi'an, China
}

\runninghead{Li, Yuan}{SymmNet, A Symmetric CNN for Occlusion Detection}

\def\etal{\emph{et al}\bmvaOneDot}

\usepackage{multirow}
\usepackage{graphicx}
\usepackage{epstopdf}
\usepackage{xfrac}
\usepackage{booktabs}
\usepackage{wrapfig}
\begin{document}

\maketitle

\begin{abstract}
Detecting the occlusion from stereo images or video frames is essential to many computer vision applications. Previous efforts focus on bundling it with the computation of disparity or optical flow, leading to a chicken-and-egg problem. In this paper, we leverage a convolutional neural network to liberate the occlusion detection task from the interleaved, traditional calculation framework. We propose a Symmetric Network (SymmNet) to directly exploit information from an image pair, without estimating disparity or motion in advance. The proposed network is structurally left-right symmetric to learn the binocular occlusion simultaneously, aimed at jointly improving both results. The extensive experiments show that our model achieves state-of-the-art results on detecting the stereo and motion occlusion.
\end{abstract}

\section{Introduction}
\label{sec:intro}

The problem of localizing the occluded and non-occluded areas over multi-view images or video sequences is of great interest for many computer vision tasks. The two  most related tasks are stereo computation and optical flow estimation. The occluded pixels violate the inter-image correspondence constraint, resulting in  ambiguous matching. State-of-the-art stereo and optical flow methods benefit from occlusion detection, either by explicitly excluding occluded pixels from disparity and motion computation \cite{alter3:conf/cvpr/BleyerRK10,Symmetric:conf/cvpr/SunLK05,occflow1:journals/pami/HeitzB93} or by repairing these  regions afterward \cite{pp1:conf/icip/HosniBGR09,pp2:journals/spl/YeGCLWZ17,MC-CNN:journals/jmlr/ZbontarL16}. Occlusion detection
also has been applied to help improve the  performance of other tasks, such as action recognition \cite{action:conf/eccv/WeinlandOF10}, object tracking \cite{track:conf/cvpr/PanH07} and 3D reconstruction \cite{3D:conf/eccv/SchonbergerZFP16}.

Most of the existing  methods take disparity or optical flow as an intermediary to estimate occlusion.
The simplest but widely used  left-right-cross-checking (LRC) \cite{LRC2:conf/eccv/TrappDH98,LRC1:journals/ijcv/HirschmullerIG02,MC-CNN:journals/jmlr/ZbontarL16} directly reasons occlusion from pre-computed disparity. This method assumes that the disparities of corresponding points in the left and right image agree with each other except for the pixels that arise from occlusion. For LRC, however, the lack of occlusion prior introduces difficulty into accurate disparity estimation. The  imperfect disparity in turn easily leads to erroneous occlusion detection, and there is no chance to revise the result.
Other approaches \cite{KZ:kolmogorov2001computing,Symmetric:conf/cvpr/SunLK05,alter1:journals/pami/YangWYSN09,alter3:conf/cvpr/BleyerRK10,alter2:conf/bmvc/VeldandiUR14} iteratively  refine their occlusion map by alternatively improving the disparity  or motion accuracy. Kolmogorov and Zabih \cite{KZ:kolmogorov2001computing} explicitly model the occlusion based on the unique matching constraint and incorporate it into an energy-based disparity estimation framework. Wang \etal \cite{wang2018occlusion} borrow the power from deep learning. They integrate a warp module for occlusion inference into an end-to-end trainable motion estimation network. This CNN-based method improves occlusion estimation coherently as learning accurate motion.
Unlike methods above deterministically deciding occlusion from disparity or optical flow, learning based method \cite{learning:conf/cvpr/HumayunAB11} uses initial motion estimations as sources to produce features for a random forest occlusion classifier. P\'{e}rez{-}R\'{u}a \etal \cite{perez2016determining} make plausible motions serve as a "soft" evidence for their occlusion model  which is based on spatio-temporal reconstruction.

To some degree, previous occlusion detectors rely on an initial estimation of disparity or optical flow. Nevertheless estimating disparity or optical flow is definitely not an easy task due to the noise, low or repetitive textures and even occlusion itself. This motivates us to explore a solution to  detect occlusion directly from stereo images or sequential frames.  In this paper, we focus on the stereo situation. Inspired by the success of convolutional neural network (CNN) in the field of monocular depth \cite{mono1:conf/nips/EigenPF14,mono2:conf/cvpr/LiuSL15,mono3:conf/cvpr/XuROWS17} and camera localization \cite{pose1:conf/iccv/KendallGC15,cameraR2:conf/cvpr/ZhouBSL17}, we leverage CNN to free occlusion detection from
disparity estimation.

We regard occlusion detection as a binary classification problem like \cite{learning:conf/cvpr/HumayunAB11} and propose a Symmetry Network (SymmNet) as the classifier. Compared with methods that infer occlusion after  regressing  the continuous disparity values or classifying disparity from hundreds of discrete labels, the high precision requirement is relaxed when directly determining the binary occlusion labels. The SymmNet is an hourglass
architecture to exploit information from binocular images. We make the network left-right symmetrically infer the binocular occlusion cooperatively, so the left and right results can be jointly improved.

The contributions of this paper are mainly three-fold:
\begin{itemize}
  \vspace{-0.8em}
  \item This work is, to the best of our knowledge, the first to directly estimate occlusion regions from images without preliminary  disparity or motion knowledge.
  \vspace{-0.8em}
  \item We propose a SymmNet which takes an image pair as input to cooperatively reason binocular occlusion.
  \vspace{-0.8em}
  \item We conduct an exhausted experimental analysis to verify our design, and our method achieves promising results for detecting stereo and motion occlusion.
\end{itemize}

\section{Proposed Model}

In binocular viewing of a scene, it is a common phenomenon that some portion of the scene can only be seen from one view.
Fig. \ref{fig:diagram} shows an example. When projecting the points in the scene onto the two views, the point $\mathbf{b}$ appears only in the
left image $I_l$ and point $\mathbf{e}$ only in the right image $I_r$. The task of pixel-wise occlusion detection is to find these monocularly visible regions given a stereo image pair. The monocularly visible regions are so-called \textbf{occlusion}.

\vspace{-1em}
\subsection{Occlusion detection with CNN}

To infer the occlusion, what information is necessary? And is CNN capable of learning it?  We argue that it is possible for CNN to learn occlusion from only one view's image in a stereo pair, while binocular images can provide more information.

\begin{wrapfigure}{r}{1.6in}
\includegraphics[width=1.6in]{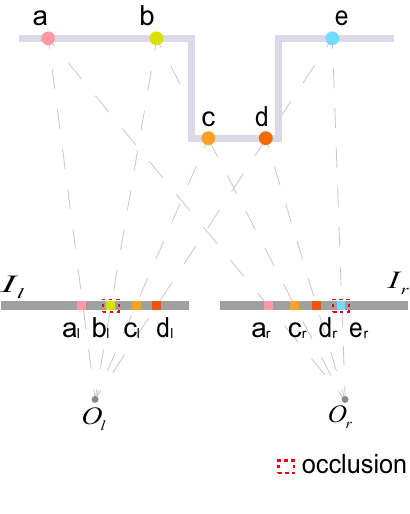}\\
\vspace{-2em}
\caption{\textbf{The occlusion reasoning diagram.} When binocular cameras $O_l$ and $O_r$ capture the scene,  $\mathbf{b}_l$ and $\mathbf{c}_r$ are occlusions on the corresponding stereo images $I_l$ and $I_r$.}\label{fig:diagram}
\vspace{-1em}
\end{wrapfigure}

\textbf{Monocular clues.}\footnote {The clues from one image in a stereo image pair, rather than an arbitrary monocular image.}  It is theoretically workable to detect occlusion by digging out monocular information. First of all, a monocular image contains depth and camera configured information, which are two basic origins of occlusion. As shown in Fig. \ref{fig:diagram}, points $\mathbf{a}$, $\mathbf{b}$, $\mathbf{c}$, and $\mathbf{d}$
are visible in image $I_l$, once we know the depth of these points as well as the pose of camera $O_r$, we can project them onto the right image $I_r$. $\mathbf{b}$ and $\mathbf{c}$ project to the same location, so the farther point $\mathbf{b}_l$ on $I_l$ can be judged as occlusion. Secondly, the local structure of an image can assist in detection, since occlusion map tends to have a specific structure corresponding to the image. For example, occlusion most likely lies just adjacent
to the edge of the closer object \cite{edgeforocc3:journals/ijcv/HoiemEH11,edgeforocc2:journals/ijcv/SteinH09} (except that the closer object is a thin stick), and the outer edge of occlusion always has the similar shape with the object edge. Furthermore, occlusion regions exhibit spatial coherence. Seldom does an isolated occluded pixel exist \cite{Symmetric:conf/cvpr/SunLK05}. Fortunately, researchers have achieved prominent results for estimating monocular depth \cite{mono1:conf/nips/EigenPF14,mono2:conf/cvpr/LiuSL15,mono3:conf/cvpr/XuROWS17}, camera pose \cite{pose1:conf/iccv/KendallGC15,pose2:conf/cvpr/KendallC17} and detecting edge \cite{edge2:conf/cvpr/BertasiusST15,edge1:journals/ijcv/XieT17} by applying CNN on these problems. This suggests that we could deal with the occlusion detection task with deep learning from a single view image.

\textbf{Binocular clues.} Although learning from a monocular image is theoretically workable, a network bears too much uncertainty to effectively encode all the necessary information including the scene geometry, camera settings, and pictorial structure.
Utilizing binocular images instead can better restrain this problem and potentially facilitate the detection in following aspects: (1) Occlusion in one image is the regions that have no correspondence in the other. Inspired by FlowNet \cite{FlowNet:conf/iccv/DosovitskiyFIHH15} which learns optical flow from two stacked frames, we consider that feeding  binocular images gives the neural network an opportunity to learn the correspondence. (2) Occlusion in one image and the depth of the other is symmetrically consistent, that is, one can trace the occlusion back to the other view's depth. As indicated in Fig. \ref{fig:diagram}, inversely project  $I_r$ to the left view according to the right depth, the being projected points ($\mathbf{a}_l$, $\mathbf{c}_l$, and $\mathbf{d}_l$) are non-occluded, otherwise (the point $\mathbf{b}_l$) is occluded.
(3) Binocular images contain the information about the relative camera pose between two views, and CNN has the ability to learn it \cite{cameraR1:conf/acivs/MelekhovYKR17,cameraR2:conf/cvpr/ZhouBSL17}.
Learning relative pose is favorable for enhancing the robustness to the changing of camera configurations.

Given a  binocular image pair, how to design an occlusion detection network? Inspired by multi-task learning \cite{multi:caruana1998multitask}, we propose to simultaneously predict the binocular occlusion. Jointly inferring the occlusion for both views is helpful to improve the prediction accuracy, since it enables consistency cross-checking between two streams. This lies in the fact that occlusion can be inferred from the depth of either view. In another word, the depth of an image is sufficient for reasoning both views' occlusion.

\begin{figure}[ht]
\centering
\includegraphics[width=4.5in]{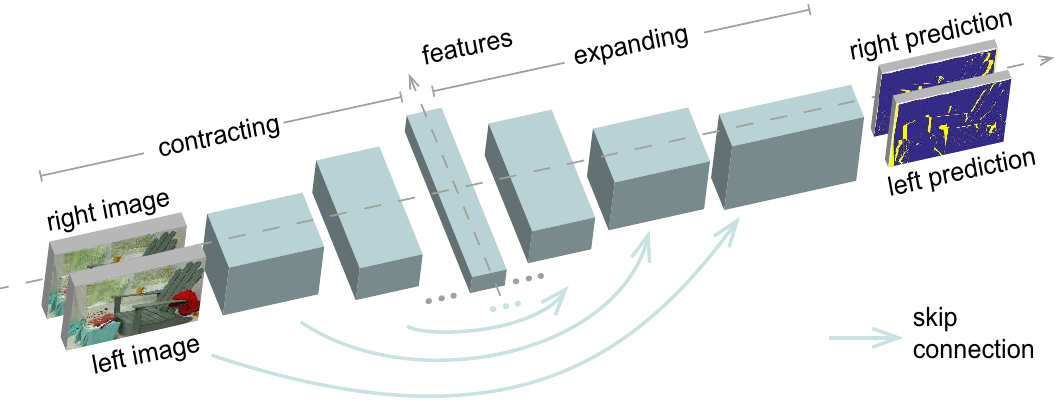}
\caption{ \textbf{SymmNet Architecture.} Intermediate layers and residual connections are omitted in the illustration.}  \label{fig:archi}
\vspace{-1.5em}
\end{figure}

Based on the observations above, we propose a \textbf{Symmetric Network (SymmNet)} which makes stacked binocular images flow through a structurally left-right symmetric neural network to predict binocular occlusion.  Fig. \ref{fig:archi} illustrates the brief architecture of the proposed network, we will introduce details in the following sections.
\label{sec:method}

\subsection{Network architecture}
\label{sec:archi}

We follow FlowNet \cite{FlowNet:conf/iccv/DosovitskiyFIHH15} to build a fully convolutional network which consists of a contractive part and an expanding part with skip connections between them. The detailed layer-by-layer definition is listed in Table \ref{table:archi}. Since determining occlusion probably relies on the information from a wide field of view, the contractive part sub-samples the features to encode large structures. It contains $6$ down-sampling layers with strides of 2 to progressively increase the receptive field and sub-samples the spatial size of feature maps by a factor of 64 in total. To obtain pixel-wise predictions with the original input resolution, in the expanding part we employ $6$ deconvolutional layers to up-sample features.  Each down- and up-sampling layer is followed with a convolutional layer for smoother results. For the sake of keeping fine local information, lower level features take part in higher level decoding through skip connections. ReLU comes after each layer to better cope with the gradient vanishing problem.

Being different from FlowNet, we take several strategies to prune the network for computational efficiency. Firstly, we compress the number of feature channels. The first layer has $16$ filters. The length of the feature is doubled every time when the spatial size of feature map is down-sampled , and reaches a maximum of 512 filters at the last layer of the contractive part. Further, we replace the concatenation in the skip connection with addition. The feature length in the expanding part is correspondingly reduced to match that in the contractive part.

Another modification is that we include an extra up-sampling module at the end of the expanding part leading to full-resolution outputs rather than half-resolution. With up-sampled to the full resolution, original image features are concatenated with the features to the last convolutional layer. This is for the consideration that low-level features from images can aid occlusion localization.

A prediction layer follows the expanding part in series to generate a 4-channel output for two views' pixel-wise occlusion classification. Every $2$ channels are normalized  as probabilities by $softmax$. Then we can get the occlusion probability  $P_L$ and $P_R$ for the left view $L$ and the right view $R$. A pixel $\mathbf{p}$ is inferred as occlusion if  $P_\cdot({\mathbf{p}})$ is larger than a threshold $\tau$.

\begin{table*}[ht]
\renewcommand\arraystretch{1.1}
\centering
\resizebox{5.1in}{!}{
\begin{tabular}{|c|ccc|c|c|c|c|ccc|c|c|}
\cline{1-6}\cline{8-13}
Name&Kernel&Str.&Ch I/O&OutRes&Input&&Name&Kernel&Str.&Ch I/O&OutRes&Input\\
\cline{1-6}\cline{8-13}
\cline{1-6}\cline{8-13}
\multicolumn{6}{|c|}{\textbf{Input}}&&\multicolumn{6}{|c|}{\textbf{Expanding}} \\
\cline{1-6}\cline{8-13}
input&&&$6/6$&$H\times W$&image pair&&
upsp5&$4\times4$&$2$&$512/256$&$\sfrac{1}{32}H\times \sfrac{1}{32}W$&conv6\\
\cline{1-6}
\multicolumn{6}{|c|}{\textbf{Contracting}}&&
iconv5&$3\times3$&$1$&$256/256$&$\sfrac{1}{32}H\times \sfrac{1}{32}W$&upsp5+conv5\\
\cline{1-6}

dwnsp1&$8\times8$&$2$&$6/16$&$\sfrac{1}{2}H\times \sfrac{1}{2}W$&input&&
upsp4&$4\times4$&$2$&$256/128$&$\sfrac{1}{16}H\times \sfrac{1}{16}W$&iconv5\\

conv1&$3\times3$&$1$&$16/16$&$\sfrac{1}{2}H\times \sfrac{1}{2}W$&dwnsp1&&
iconv4&$3\times3$&$1$&$128/128$&$\sfrac{1}{16}H\times \sfrac{1}{16}W$&upsp4+conv4\\

dwnsp2&$6\times6$&$2$&$16/32$&$\sfrac{1}{4}H\times \sfrac{1}{4}W$&conv1&&
upsp3&$4\times4$&$2$&$128/64$&$\sfrac{1}{8}H\times \sfrac{1}{8}W$&iconv4\\

conv2&$3\times3$&$1$&$32/32$&$\sfrac{1}{4}H\times \sfrac{1}{4}W$&dwnsp2&&
iconv3&$3\times3$&$1$&$64/64$&$\sfrac{1}{8}H\times \sfrac{1}{8}W$&upsp3+conv3\\

dwnsp3&$6\times6$&$2$&$32/64$&$\sfrac{1}{8}H\times \sfrac{1}{8}W$&conv2&&
upsp2&$4\times4$&$2$&$64/32$&$\sfrac{1}{4}H\times \sfrac{1}{4}W$&iconv3\\

conv3&$3\times3$&$1$&$64/64$&$\sfrac{1}{8}H\times \sfrac{1}{8}W$&dwnsp3&&
iconv2&$3\times3$&$1$&$32/32$&$\sfrac{1}{4}H\times \sfrac{1}{4}W$&upsp2+conv2\\

dwnsp4&$4\times4$&$2$&$64/128$&$\sfrac{1}{16}H\times \sfrac{1}{16}W$&conv3&&
upsp1&$4\times4$&$2$&$32/16$&$\sfrac{1}{2}H\times \sfrac{1}{2}W$&iconv2\\

conv4&$3\times3$&$1$&$128/128$&$\sfrac{1}{16}H\times \sfrac{1}{16}W$&dwnsp4&&
iconv1&$3\times3$&$1$&$16/16$&$\sfrac{1}{2}H\times \sfrac{1}{2}W$&upsp1+conv1\\

dwnsp5&$4\times4$&$2$&$128/256$&$\sfrac{1}{32}H\times \sfrac{1}{32}W$&conv4&&
upsp0&$4\times4$&$2$&$16/8$&$H\times W$&iconv1\\

conv5&$3\times3$&$1$&$256/256$&$\sfrac{1}{32}H\times \sfrac{1}{32}W$&dwnsp5&&
iconv0&$3\times3$&$1$&$14/8$&$H\times W$&upsp0 $\bigoplus$ input\\
\cline{8-13}
dwnsp6&$4\times4$&$2$&$256/512$&$\sfrac{1}{64}H\times \sfrac{1}{64}W$&conv5&&
\multicolumn{6}{|c|}{\textbf{Prediction}}\\
\cline{8-13}
conv6&$3\times3$&$1$&$512/512$&$\sfrac{1}{64}H\times \sfrac{1}{64}W$&dwnsp6&&
pr&$3\times3$&$1$&$8/4$&$H\times W$&iconv0\\
\cline{1-6}\cline{8-13}

\end{tabular}}
\vspace{-1em}
\caption{\textbf{SymmNet architecture summary.} Each layer except for the prediction layer \textit{pr} is followed by ReLU. \textit{pr} layer is followed by \textit{softmax} to generate probability. This table is arranged from top to bottom , left to right. $+$ is the addition operation, $\bigoplus$ is the concatenation operation in skip connection.}
\vspace{-1em}
\label{table:archi}
\end{table*}

\vspace{-0.5em}
\subsection{Training details}
\label{sec:train}
To jointly train the binocular occlusion, we use the total binary-cross-entropy loss of both views as objective:
\begin{equation}
\begin{split}
\mathbf{L} = -\frac{1}{2}&\left( w_L^o\sum_{\mathbf{p}}\mathbf{1}(O_L(\mathbf{p})=1)\log(P_L(\mathbf{p}))+
w_L^{\bar{o}}\sum_{\mathbf{p}}\mathbf{1}(O_L(\mathbf{p})=0)\log(1-P_L(\mathbf{p}))\right.\\
&\left.+ w_R^o\sum_{\mathbf{p}}\mathbf{1}(O_R(\mathbf{p})=1)\log(P_R(\mathbf{p}))+
w_R^{\bar{o}}\sum_{\mathbf{p}}\mathbf{1}(O_R(\mathbf{p})=0)\log(1-P_R(\mathbf{p}))\right),\\
\end{split}
\end{equation}
where $O_{\cdot}$ is ground-truth occlusion, $\mathbf{1}(\cdot)$ is indicating function, $w^c_{\cdot}$ is a class weight to make the loss adapt to the unbalanced number of occlusion and non-occlusion pixels.   We adopt the bounded class weight \cite{BW:journals/corr/PaszkeCKC16} $w^c_{\cdot} = {1}/{\ln (\epsilon + q^c_{\cdot})}$, where $q^c_{\cdot}$ is the proportion of class $c$ (occlusion $o$ or non-occlusion $\bar{o}$) in the training batch. $\epsilon$ is a hyper-parameter to limit the weight range.

We trained our model on the SceneFlow dataset \cite{SF:conf/cvpr/MayerIHFCDB16}, which consists of stereo image pairs rendered from synthetic sequences. The dataset is suitable for training the network for two reasons. One is that this dataset contains 35, 454 training and 4, 370 test pairs. It is large enough to train the model without over-fitting. The other reason is that it provides dense, perfect ground-truth disparity for both views, which can be used to generate binocular ground-truth occlusion.  The ground-truth  $O_v$ for a view $v$ is obtained by checking the left-right-consistency between its ground-truth disparity $D_v$ and the other view's, as
\begin{equation}
\label{eqn:gt}
O_v(\mathbf{p}) = \mathbf{1}\left(|D_{v}(\mathbf{p})-\hat{D}_{v}(\mathbf{p})|>\delta\; \right),\quad v \in \{L,R\}.
\end{equation}
$\hat{D}_{v}$ is the warped disparity from the other view $v'$. It is obtained by bilinear sampling mechanism \cite{warp:conf/nips/JaderbergSZK15} as $\hat{D}_{v}(\mathbf{p}) = \sum_{i\in \{t,b\}, j \in \{l,r\}}\omega^{ij} D_{v'}(\mathbf{t}^{ij})$. $\mathbf{t}^{ij}$ is the 4-pixel neighbors of $\mathbf{t}$, which is the corresponding position of $\mathbf{p}$ on view $v'$  based on $D_v(\mathbf{p})$. $\omega_{ij}$ is the interpolation weight and $\sum_{i,j}\omega_{i,j}=1$.

Training samples are randomly cropped patches with a spatial size of $H = 256$ and $W = 768$.
The cropping process is for computational restriction. Besides, it is a data augmentation means, since the shape of the out-of-image occlusion at image boundary varies as cropping a patch at different locations. Accordingly, it should be noted that the ground-truth computation has to be done after cropping due to the varying out-of-image occlusion.

The network were optimized using the Adam \cite{adam:journals/corr/KingmaB14} method ($\beta_1 = 0.9$ and $\beta_2 = 0.99$) and a constant learning rate of $1 \times 10^{-2}$ for $10$ epoches. The training batch contains $16$ samples. $\varepsilon$ in the class weight is empirically set to $1.5$, $\delta$ in Eq. (\ref{eqn:gt}) is set to 1.

\section{Experiment}
\label{sec:expe}
In this section, we first test several variants of our method to verify the proposed pipeline.
Then we compare the overall performance with several existing methods on SceneFlow \cite{SF:conf/cvpr/MayerIHFCDB16} and Middlebury \cite{MB02:journals/ijcv/ScharsteinS02,MB03:conf/cvpr/ScharsteinS03,MB06:conf/cvpr/HirschmullerS07,MB05:journals/ijcv/PalWTS12,MB:conf/dagm/ScharsteinHKKNWW14} dataset. Furthermore, we examine our model's capacity to learn motion occlusion
on MPI Sintel dataset \cite{MPI:journals/tip/JacobsonFN12}. We finally report the time and memory requirement of our architecture.

\begin{figure}[t]
\centering
\includegraphics[width=4.8in]{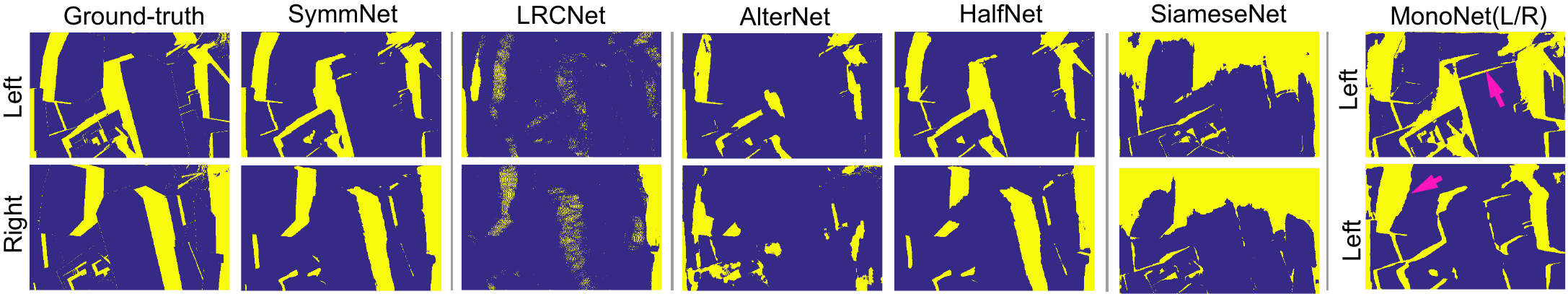}
\caption{\textbf{Example results of different architecture variants.} The first row in MonoNet(L/R) column is the result of MonoNetL, the second row is MonoNetR. The pink arrow in MonoNetL points to the fake occlusion occurring at the image edge. The arrow in MonoNetR points to the erroneous occlusion shape.}  \label{fig:ablation}
\vspace{-1em}
\end{figure}

For evaluation, we report three metrics commonly used in occlusion detection task, which are precision (the percentage of true occluded pixels in  detected occlusion), recall (the percentage of the detected occlusion pixels in the occluded regions) and Fscore (the harmonic average of precision and recall). When predicting occlusion, the threshold $\tau$ is set to $0.5$ unless otherwise specified.

\vspace{-1em}
\subsection{Architecture Analysis}
\label{sec:Ablation}

\begin{wrapfigure}{R}{1.7in}
\vspace{-0.8em}
\includegraphics[width=1.7in]{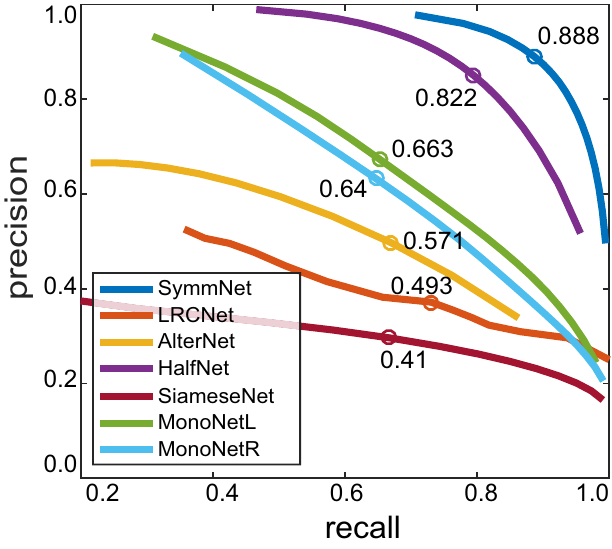}\\
\vspace{-2em}
\caption{\textbf{Precision-recall (PR) curves.} The max Fscore on a PR curve is annotated. The curve that is closer to the upper-right-hand corner is better. SymmNet outperforms all of its variants.}\label{fig:PR}
\vspace{-1em}
\end{wrapfigure}

To justify our design choices, we test several model variants of \textbf{SymmNet} on the SceneFlow test set. To be fair, we keep the parameter number of different architectures all the same except for the input and output layers.
 In Fig. \ref{fig:ablation}, we provide example results and in Fig. \ref{fig:PR} we visualize the precision-recall (PR) curves.

%
%
%
%

\textbf{MonoNet(L/R).} To investigate the role of monocular image input, we modify the SymmNet to take the single left image and right image separately as input and to predict the left occlusion (MonoNetL and MonoNetR). Fig. \ref{fig:ablation} shows that either image serves occlusion detection but in a different manner.  The image of the homogeneous view, i.e., the left image, tends to provide more information about the object edges. This is a useful clue for determining the shape of occlusion, while gives rise to fake occlusion. The image of the cross view helps to tell the true occluded edges, while performs poor at estimating the shape.

\textbf{SiameseNet.} A Siamese architecture comprises two sub-networks with shared weights \cite{Siamese}. This architecture is widely adopted in the highly related task of stereo matching, in which each of the two branches concentrates on one view (the left or right) to extract unary features \cite{MC-CNN:journals/jmlr/ZbontarL16,content:conf/cvpr/LuoSU16} or to regularize the cost volume \cite{GCnet:conf/iccv/KendallMDH17}. The Siamese structure, treating the binocular views indiscriminately, is efficient for stereo methods  to encode the shared-knowledge  as well as reduce the computational requirement. For occlusion detection, we construct a similar Siamese variant that each branch takes one view as input and outputs the corresponding occlusion map. The disordered results in Fig. \ref{fig:ablation} shows its incompetency for occlusion detection. This is because reasoning occlusion is subject to the viewpoint and we are aiming at finding the view-subjected discrimination. The shared-weights disable the network to learn the distinctive information. In addition,  seeing one view for each shared-branch enforces the network only to involve monocular cues while omit the vital binocular information, such as the relative pose of the fellow camera.

\textbf{AlterNet \& HalfNet.} Our model jointly learns binocular occlusion in order to make two streams help each other learn better. To verify this design, we construct two variants based on SymmNet for comparison.
One is AlterNet that only outputs occlusion for a single view, while we iteratively interchange the stacked order of two input images to alternatively learn either the left or right occlusion as training. The other is HalfNet which consists of two separate networks, one for learning each view's occlusion independently. Each sub-network in the HalfNet still takes binocular images as input, but the length of feature channel is half of that in SymmNet so as to keep the total model volume unchanged. AlterNet gets into trouble when  learning the alternating views. As shown in Fig. \ref{fig:ablation}, the result of the left view is approximately correct, while the right result gets a mess. HalfNet equally estimates both occlusion maps with good quality, while numerically performs slightly worse than SymmNet.

\textbf{LRCNet.} An alternative method to detect occlusion is accurately estimating disparity first and inferring occlusion from disparity instead. We replace the prediction layer in SymmNet with a regression layer to make the network learn binocular disparity and then apply LRC on the disparity to infer occlusion. We call this network LRCNet. This network lacks a module to directly regularize the shape of occlusion, thus there are evident holes in the occlusion regions as shown  in Fig. \ref{fig:ablation}. Moreover, the disparity results directly determine the occlusion detection quality, while learning disparity seems not easy. Among the variants, this network is the only one that is used for learning disparity rather than occlusion, whereas its performance is rather poor.

\textbf{Discussion.} Our SymmNet directly models occlusion from input images, rather than inferring occlusion at the following stage of disparity computation. This design, on the one hand, eases the problem in terms of engineering, as can be seen from the significant gap between the PR curves of LRCNet and ours in Fig. \ref{fig:PR}. On the other hand, it can be integrated into the disparity estimation framework at the very beginning, as suggested by Anderson and Nakayama that one senses occlusion at the earliest stages in the binocular visual system \cite{bg1:anderson1994toward}.

We learn complementary information from binocular images. Both images are indispensable for precise estimation, particularly for eliminating the fake occlusion and keeping the shape of occlusion. More importantly, two images work together, providing relative cues which are necessary for determining the discrimination between two views. This can be verified by comparing the results of SymmNet with MonoNets and SiameseNet in Fig. \ref{fig:ablation}.

Furthermore, it should be noted that although we follow FlowNet to construct our structurally similar architecture, these two networks are different in nature. FlowNet only predicts for the reference view, while our SymmNet uses a unified architecture to reason the occlusion for both views. This symmetric design enables knowledge to transfer between the two views and boosts their performance coherently.  The PR curve of our model covers the curves of all the variants from the upper-right, which shows the reasonability of our whole design.

\begin{figure}[t]
\centering
\includegraphics[width=4.8in]{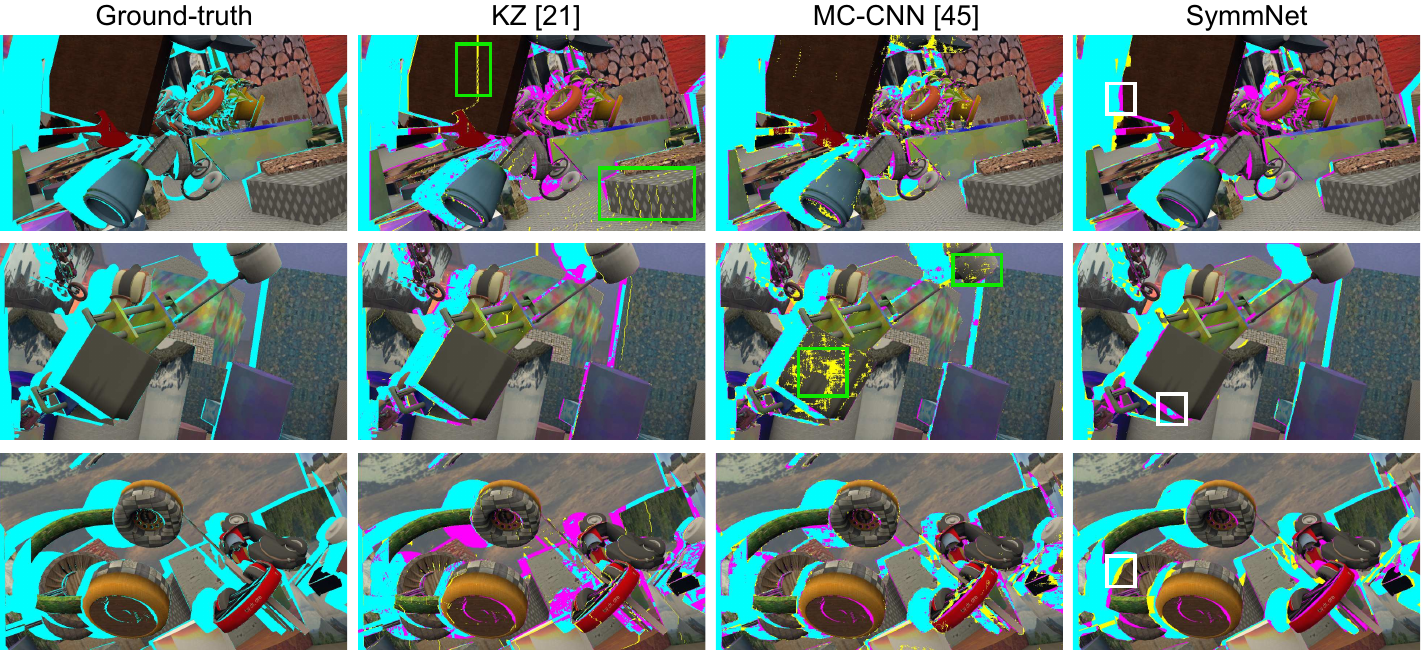}
\caption{\textbf{Qualitative comparison on SceneFlow dataset.} The true positive estimations are labelled in cyan, the false negative in magenta and the false positive in yellow. The green boxes in the first row label the errors occur at slanted planes, and the errors at textureless regions in the second row. The white boxes in the last column show the failure cases of our method.}  \label{fig:SF}
\vspace{-1.5em}
\end{figure}

\vspace{-1.0em}
\subsection{Overall Performance}
\label{sec:Overall}
We compare our overall performance with that of two other occlusion detectors. We first run the method of Kolmogorov and Zabih \cite{KZ:kolmogorov2001computing} (KZ) which enforces the uniqueness constraint to detect the un-matched pixels as occlusion. We also compare with the LRC method. The initial disparity for LRC is obtained as MC-CNN \cite{MC-CNN:journals/jmlr/ZbontarL16}, i.e.,  by extracting and matching the deep features, followed cross-based cost aggregation \cite{cross:journals/tcsv/ZhangLL09} and semiglobal matching \cite{sgm:journals/pami/Hirschmuller08}. We use
the code provided by the authors of these methods.
\vspace{-1.0em}
\subsubsection*{Validation on SceneFlow}
\vspace{-0.5em}
We first evaluate the performance on SceneFlow test set. For a fair comparison, we fine-tune the MC-CNN model on the SceneFlow training set and report the best scores we can get.
Fig. \ref{fig:SF} exhibits the qualitative results. The performance of KZ and LRC relies on the quality of the initial disparity. KZ fails to recover the disparity of the slanted plane due to the first-order smoothness prior,  MC-CNN encounters difficulty at matching the large textureless regions. Consequently, the occlusion detection error easily appears in the corresponding regions as shown in the first  two rows in Fig. \ref{fig:SF}. Our method, directly predicting the occlusion regions, is free from the influence of the initial disparity estimation. Quantitative results also show the superiority of our method as summarized in Table \ref{table:scores}.

Although our method obtains largely proper results, we fail to make the occlusion precisely coincide with the image edges, as shown in Fig. \ref{fig:SF}. We ascribe this failure to the smoothing effect brought by the convolution and contraction operations in the proposed architecture. Trying to rethink an architecture to preserve more details may help alleviate the problem. In addition, explicit matching evidence, rather than our implicit matching cues from stacked images, is potentially beneficial to accurate location.

\begin{figure}[t]
\centering
\includegraphics[width=4.8in]{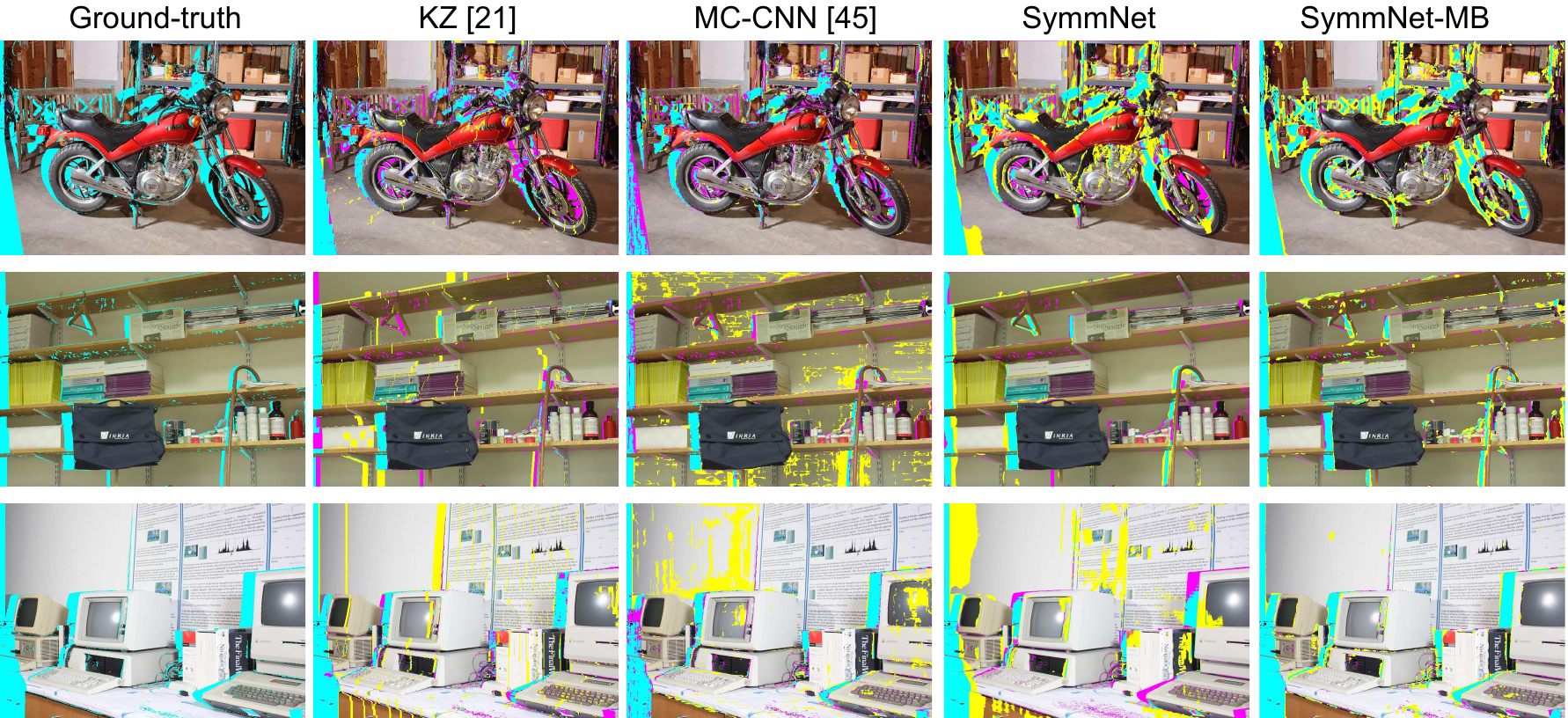}
\caption{\textbf{Qualitative comparison on Middlebury dataset.}  The true positive estimations are
labelled in cyan, the false negative in magenta and the false positive in yellow.}  \label{fig:MB}
\vspace{-1.0em}
\end{figure}
\vspace{-1.0em}
\subsubsection*{Validation on Middlebury}
\vspace{-0.5em}
\label{sec:MB}
Middlebury dataset provides stereo image pairs with dense ground-truth disparity of indoor scenes under controlled lighting conditions.
Compared to the SceneFlow dataset, the scenes are more realistic, the lighting conditions and exposure settings are more complex. We collect 2845 image pairs with ground-truth disparity of both views, and split the collections into training set and validation set to conduct 10-fold-cross-validation. For our method, we test two configurations: (1) directly applying the model trained on SceneFlow and (2) fine-tuning the model for another 50 epoches on the Middlebury training set with learning rate set to $1 \times 10^{-3}$ (SymmNet-MB). $\varepsilon$ in the class weight is adjusted to $1.2$ for the smaller occlusion.

The qualitative results are shown in Fig. \ref{fig:MB} and quantitative results are given in Table \ref{table:scores}. Our fine-tuned model outperforms other methods on all the evaluation indexes. It is worth noting that our method reveals some robustness to the variation of camera configurations and environment. Since even without fine-tuning,  our method can also generate comparable results.


\label{sec:MPI}
\begin{table*}[t]
\renewcommand\arraystretch{1.1}
\centering
\resizebox{5.1in}{!}{
\begin{tabular}{|l|c|c|c|c|c|c|l|c|c|}
\hline
\multicolumn{7}{|c|}{Stereo Occlusion}& \multicolumn{3}{|c|}{Motion Occlusion} \\
\hline\hline
&\multicolumn{3}{|c|}{SceneFlow}&\multicolumn{3}{|c|}{Middlebury}& \multicolumn{3}{|c|}{MPI} \\
\hline
&Precision&Recall&F-score&Precision&Recall&F-score&&Oracle 69&Global 69\\
\hline
KZ\cite{KZ:kolmogorov2001computing}&0.554&0.609&0.580&0.585&0.628&0.605&Learning \cite{learning:conf/cvpr/HumayunAB11}&0.535&0.448\\
MC-CNN\cite{MC-CNN:journals/jmlr/ZbontarL16}$+$LRC&0.772&0.836&0.802&0.660&0.664&0.652&Depth Order \cite{depthorder:conf/nips/SunSB10}&0.465&0.449\\
SymmNet&\textbf{0.799}&\textbf{0.919}&\textbf{0.873}&0.584&0.737&0.666&P\'{e}rez{-}R\'{u}a \etal \cite{perez2016determining}&0.550&0.540\\
SymmNet-MB&-&-&-&\textbf{0.810}&\textbf{0.849}&\textbf{0.828}&Ours-MPI&\textbf{0.665}&\textbf{0.642}\\
\hline
\end{tabular}}
\vspace{-0.5em}
\caption{\textbf{Quantitative evaluations.} All the evaluations are the higher, the better. We highlight the best scores in bold.}
\label{table:scores}
\end{table*}

\begin{figure}[t]
\centering
\includegraphics[width=4.8in]{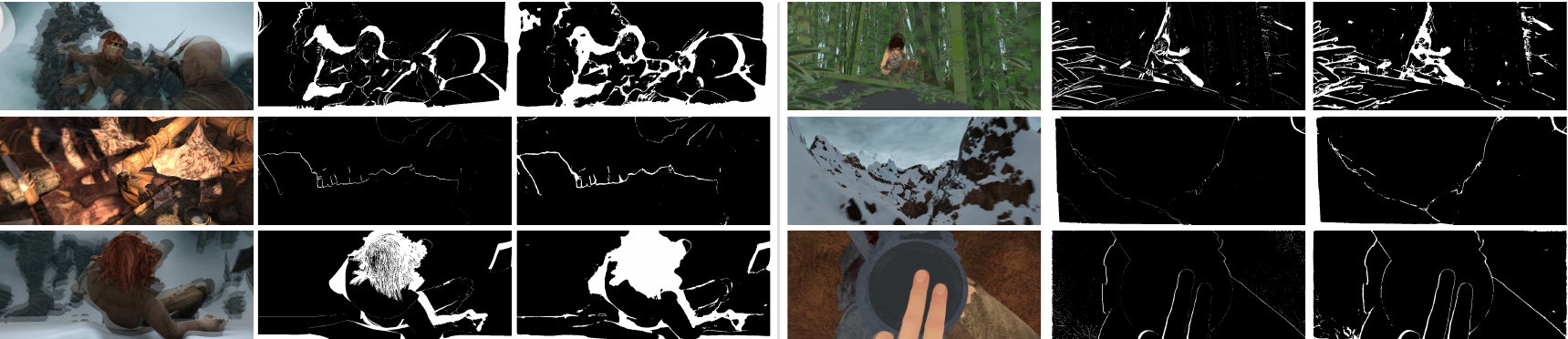}
\vspace{-0.5em}
\caption{\textbf{Qualitative results of our method on MPI dataset.} From left to right:  Average image of the two input frames; Occlusion ground-truth; Predicted occlusion map with proposed method.}  \label{fig:MPI}
\vspace{-1.5em}
\end{figure}
\vspace{-1.0em}
\subsection{Motion occlusion detection}
Although our model is designed for detecting the occlusion in stereo settings, it can be effortlessly adapted to the task of motion occlusion detection by taking two consecutive frames as inputs. We demonstrate this ability on the MPI Sintel dataset.  This dataset contains 69 sequences (3123 image pairs) equipped with ground-truth occlusion maps. We divide the dataset into training set and validation set for 10-fold-cross-validation. The hyper parameter $c$ is set to $1.01$ to fit the
extremely unbalanced occlusion ratio in this experiment.

We compare with three motion occlusion detectors: the learning based method \cite{learning:conf/cvpr/HumayunAB11}, a depth order based method \cite{depthorder:conf/nips/SunSB10} and the spatial-temporal reconstruction model of  P\'{e}rez{-}R\'{u}a \etal \cite{perez2016determining}. Following the evaluation methodology of P\'{e}rez{-}R\'{u}a \etal, we test the average F-score over all 69 sequences when the threshold $\tau$ is set to maximize F-score (Oracle 69) and to $0.5$ (Global 69). Our method excels all the other methods on both settings as shown in Table \ref{table:scores}.

We provide several detection results in Fig. \ref{fig:MPI}. Even though the true occlusion regions are much smaller and finer than those in stereo, we can still make a good prediction.

\vspace{-1.0em}
\subsection{Runtime and memory requirement}
We test the runtime of our PyTorch implementation on a single NVIDIA Tesla M40 GPU. Training on SceneFlow dataset can be finished in two days. It takes 0.07s and requires 651M graphic memory to predict an image pair in size of $540 \times 960$. The low requirement of time and memory makes our model an optional preprocess module for other tasks such as object tracking, human pose estimation and action recognition.

\vspace{-1.0em}
\section{Conclusion}
\label{sec:conclusion}

We have proposed a CNN model called SymmNet to detect occlusion from stereo images or video sequences. Unlike the  traditional occlusion detectors which infer occluded pixels from pre-computed disparity or optical flow, our model directly learns from original images. The proposed SymmNet is left-right symmetric to jointly learn binocular occlusion by cooperatively extracting the binocular information. The experiment results have demonstrated the good ability of our method for stereo and motion occlusion detection.

We believe the proposed occlusion detector can be extended to facilitate other applications, such as stereo and optical flow.
It would be an interesting future work to investigate the auxiliary role of occlusion based on our method.

\vspace{1em}
\noindent{\textbf{Acknowledgement:}} This work was supported by the National Key R$\&$D Program of China (No.2016YFB1001001) and the National Natural Science Foundation of China (No.61573280, No.91648121).

\bibliography{OccNet_final}
\end{document}